\documentclass[runningheads]{llncs}
 
\usepackage[T1]{fontenc}
\usepackage{amsfonts}
\usepackage{amsmath}
\usepackage{amssymb}
\usepackage{graphicx}
\usepackage{comment}
\usepackage{epstopdf}
\usepackage{algorithmic}
\usepackage{hyperref}
\usepackage{url}
 
\ifpdf
  \DeclareGraphicsExtensions{.eps,.pdf,.png,.jpg}
\else
  \DeclareGraphicsExtensions{.eps}
\fi
 
\usepackage{amsopn}

\begin{document}
 
\title{Rashomon Alignment}
 
\titlerunning{Rashomon Alignment}
 
\author{Moisés Santos\inst{1,2} \and Peter van der Putten\inst{5} \and Bernhard Pfahringer\inst{6} \and Carlos Soares\inst{1,2,3,4}} 
\authorrunning{Santos et al.}

\institute{Faculty of Engineering, University of Porto, Portugal \and
Artificial Intelligence and Computer Science Lab (LIACC), Portugal \and
Fraunhofer AICOS Portugal, Portugal \and
BrightFactory, Portugal \and
LIACS, Leiden University, Leiden, The Netherlands \and
School of Computing and Mathematical Sciences, University of Waikato, Hamilton, New Zealand \\
\email{mrsantos@fe.up.pt, p.w.h.van.der.putten@liacs.leidenuniv.nl, bernhard@cs.waikato.ac.nz, csoares@fe.up.pt}
}

\maketitle
 
\begin{abstract}
We propose Rashomon Alignment (RA), a new measure to assess functional similarity between two models. Existing functional similarity measures are \emph{distributional}, quantifying differences between outputs of models applied to real-world data.
However, these measures can be regarded as ecologically valid only for regions in the input space represented by the available data.
We introduce a \emph{geometrical} perspective on functional model similarity, which estimates it across the entire data space, offering a comprehensive view of decision boundary alignment independent of any specific data distribution. We also propose geometric Rashomon Alignment as a measure of geometrical similarity, which is computed using data uniformly sampled from the instance space. We perform an experimental analysis on more than 90 datasets, examining critical cases where model alignment diverges from predictive accuracy.
Our results show that geometrical and distributional alignment provide different and complementary perspectives on the similarity between models and algorithms. RA can be used for multiple purposes, including model selection, ensemble construction, and enhanced interpretability of machine learning models and algorithms.
\end{abstract}
 

\section{Introduction}
 
Functional measures assess the similarity between models by comparing their outputs and have been used for multiple purposes, including a better understanding of models and algorithms, improve learning and select models in ensemble learning~\cite{Klabunde2025}. However, current usage of these measures consists of applying them to the outputs of the models being evaluated on available data~\cite{Lee2011,Klabunde2025}. This means that the representativeness of the available data affects the quality of the estimates of similarity obtained this way. Poorly-representative samples lead to poor estimations of similarity between models. Additionally, even if the estimates are accurate with the current data, they will become outdated in the case of concept change.

To address this question, we propose \emph{Rashomon%
\footnote{While Rashomon Sets originally describe sets of models with high predictive performance~\cite{Breiman2001,Fisher2019}, the idea of similarity between models is as interesting as the focus on high-accuracy models. Therefore, we believe it adequate to use this as the name of a measure that can be used to identify models that define similar decision boundaries.} Alignment}%
\footnote{We prefer the term \emph{alignment} to \emph{similarity} as two models can be very different (i.e.\ the descriptions of the same concept they represent are very different) but define similar decision boundaries in the instance space}
(RA).
Since it is neither possible to assess the representativeness of the available data sample regarding the corresponding true distribution, nor to sample from that distribution, as it is unknown, we propose the use of a representative sample of the data space (i.e., the space defined by the domains of the variables). We refer to this as \emph{Geometrical Rashomon Alignment} (gRA), as opposed to \emph{Distributional Rashomon Alignment} (dRA), which would be obtained using a sample that is representative of the distribution of the data.

We illustrate the proposed measures on the comparison between pruned and unpruned decision trees~\cite{Aggarwal2015}. The results are obtained on 92 UCI datasets~\cite{UCI2025}.
We show that the geometric and distributional alignment provide different perspectives on the similarity between models.
We also show that, as expected, the method is able to identify cases when the two models are in fact producing quite different classifications but have similar accuracy, which demonstrates that just focusing on predictive performance alone can be deceiving.
 
The contributions in this paper are:
\begin{itemize}
\item a new method to assess the similarity between classification models and algorithms;
\item an extensive empirical study of the new method, to illustrate its potential.
\end{itemize}
 
The remainder of this paper is organized as follows. In Sect.~\ref{sec:align} we discuss alignment of algorithms in terms of produced classifications in more detail, and relate it to predictive performance, as well as performance on synthetic versus real world data. We then discuss our experimental setup and results in respectively Sects.~\ref{sec:setup} and \ref{sec:results}, and the conclusions in Sect.~\ref{sec:conclusion}.

\section{Rashomon Alignment}
\label{sec:align}

Empirical comparisons between learning algorithms typically rely on predictive performance, but accuracy alone does not reveal whether two models err in the same regions of the instance space or define structurally similar decision boundaries~\cite{Fernandez-Delgado2014a,d2022}. \emph{Functional similarity measures}~\cite{Klabunde2025} address this gap by comparing model outputs rather than internal representations. We build on the prediction-based variant, which we now formalize and extend with a geometric perspective.

\subsection{Prediction-Based Agreement}
\label{sec:agreement}

Let $\mathcal{C}$ be a finite set of class labels and $\mathcal{F} = \mathcal{F}_1 \times \cdots \times \mathcal{F}_p$ the instance space, where each $\mathcal{F}_k$ is the domain of attribute $k$. Let $M_A, M_B : \mathcal{F} \to \mathcal{C}$ be two models induced by algorithms $A$ and $B$ on training data drawn from a distribution $P$ over $\mathcal{F} \times \mathcal{C}$ with marginal $P(X)$ on $\mathcal{F}$.

For any finite set of evaluation instances $X = \{x_1, \ldots, x_n\} \subseteq \mathcal{F}$, the empirical \emph{agreement} between $M_A$ and $M_B$ on $X$ is:%
\footnote{Equivalent to the complement of the \emph{disagreement} distance~\cite{Lee2011}, and related to the Jaccard coefficient~\cite{jaccard1901}. Soft variants based on Cohen's Kappa~\cite{cohen1960} or predicted probabilities are equally applicable.}
\begin{equation}
  \label{eq:agreement}
  \theta_X(M_A, M_B) = \frac{1}{n} \sum_{i=1}^{n} \mathbf{1}\bigl[M_A(x_i) = M_B(x_i)\bigr] \in [0, 1].
\end{equation}
A value of $1$ indicates identical predictions on $X$, $0$ indicates full disagreement. The empirical agreement is an estimator of the population-level agreement $\theta^Q(M_A, M_B) = \mathbb{E}_{x \sim Q}[\mathbf{1}[M_A(x) = M_B(x)]]$ with respect to some reference distribution $Q$ over $\mathcal{F}$. The choice of $Q$ determines what notion of similarity is being measured, and it is precisely this choice that distinguishes the two variants of RA introduced below. We call both variants \emph{Rashomon Alignment} since they identify pairs of models with equivalent predictive behavior~\cite{d2022,Breiman2001}.

\subsection{Distributional Rashomon Alignment}
\label{sec:dra}

dRA takes the reference distribution $Q$ to be the data-generating distribution $P(X)$. Formally:
\begin{equation}
  \label{eq:dra-pop}
  \mathrm{dRA}^*(M_A, M_B)
  = \mathbb{E}_{x \sim P(X)}\bigl[\mathbf{1}[M_A(x) = M_B(x)]\bigr]
  = \int_{\mathcal{F}} \mathbf{1}[M_A(x) = M_B(x)]\, p(x)\, dx,
\end{equation}
where $p(x)$ denotes the density of $P(X)$. Given a sample $X_\text{obs} = \{x_1, \ldots, x_n\}$ drawn i.i.d.\ from $P(X)$ (in practice, a hold-out test set), the dRA is estimated by:
\begin{equation}
  \label{eq:dra}
  \mathrm{dRA}(M_A, M_B) = \theta_{X_\text{obs}}(M_A, M_B),
\end{equation}
which is an unbiased and consistent estimator of $\mathrm{dRA}^*$ by the law of large numbers, with $\mathrm{dRA} \xrightarrow{a.s.} \mathrm{dRA}^*$ as $n \to \infty$.

The dRA reflects agreement in the regions of $\mathcal{F}$ where real data lies. However, it inherits the representativeness of $X_\text{obs}$ and is sensitive to changes in $P(X)$: under covariate shift, the same pair of models may yield arbitrarily different dRA values without their decision boundaries having changed.

\subsection{Geometric Rashomon Alignment}
\label{sec:gra}

gRA takes the reference distribution $Q$ to be the uniform distribution $U(\mathcal{F})$ over the instance space, so that all regions of $\mathcal{F}$ contribute equally regardless of where data is observed:
\begin{equation}
  \label{eq:gra-pop}
  \mathrm{gRA}^*(M_A, M_B)
  = \mathbb{E}_{x \sim U(\mathcal{F})}\bigl[\mathbf{1}[M_A(x) = M_B(x)]\bigr]
  = \frac{1}{|\mathcal{F}|} \int_{\mathcal{F}} \mathbf{1}[M_A(x) = M_B(x)]\, dx,
\end{equation}
where $|\mathcal{F}| = \int_\mathcal{F} dx$ is the volume of the instance space (assumed bounded; for unbounded domains, $\mathcal{F}$ is restricted to the empirical range observed in training data). Geometrically, $\mathrm{gRA}^*$ is the fraction of the instance space on which the two decision functions coincide.

Given a synthetic sample $X_\text{syn} = \{x_1^\star, \ldots, x_m^\star\}$ drawn i.i.d.\ from $U(\mathcal{F})$, the gRA is estimated by:
\begin{equation}
  \label{eq:gra}
  \mathrm{gRA}(M_A, M_B) = \theta_{X_\text{syn}}(M_A, M_B),
\end{equation}
which is again unbiased and consistent: $\mathrm{gRA} \xrightarrow{a.s.} \mathrm{gRA}^*$ as $m \to \infty$. Crucially, $X_\text{syn}$ is generated independently of any training or test data, so $\mathrm{gRA}$ is invariant to changes in $P(X)$ as long as the underlying decision functions $M_A$ and $M_B$ remain fixed.

\subsection{Relation Between dRA and gRA}
\label{sec:variants}

\begin{figure}[tbp]
\centering
      \includegraphics[width=0.7\textwidth]{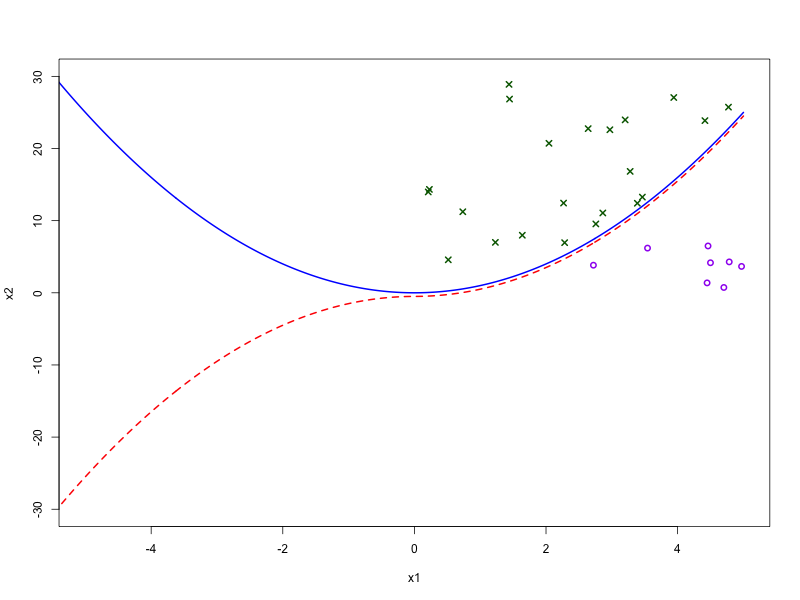}
      \caption{Two decision boundaries with high distributional but low geometric alignment.}
      \label{fig:dif-dist-geom-align}
\end{figure}

The two variants are related through a change of measure. Writing the dRA as an expectation under the uniform distribution via the importance ratio $|\mathcal{F}|\, p(x)$ yields:
\begin{equation}
  \label{eq:relation}
  \mathrm{dRA}^*(M_A, M_B) = |\mathcal{F}| \int_{\mathcal{F}} \mathbf{1}[M_A(x) = M_B(x)]\, p(x)\, dx,
\end{equation}
so that $\mathrm{dRA}^* = \mathrm{gRA}^*$ when $p(x) = 1/|\mathcal{F}|$ (uniform $P(X)$), and the two diverge otherwise. The magnitude of the gap $|\mathrm{dRA}^* - \mathrm{gRA}^*|$ is bounded above by the total variation between $P(X)$ and $U(\mathcal{F})$, and quantifies the extent to which observed data fails to represent the full instance space. This gap is what makes gRA informative beyond dRA: a high dRA with low gRA indicates that two models agree where data lies but disagree elsewhere, a form of agreement that is fragile under distribution shift. Figure~\ref{fig:dif-dist-geom-align} illustrates such a case.

Both measures extend to algorithm-level comparison by averaging over a collection of datasets $\mathcal{X} = \{X_1, \ldots, X_N\}$:
\[
  \theta_\mathcal{X}(A, B) = \frac{1}{N} \sum_{i=1}^{N} \theta_{X_i}(M_{A,i}, M_{B,i}),
\]
where $M_{A,i}$ and $M_{B,i}$ are the models trained by $A$ and $B$ on dataset $i$.

\subsection{Alignment vs.\ Accuracy}
\label{sec:disc-alig-vs-acc}

\begin{figure}[tbp]
  \centering
  \includegraphics[width=0.85\textwidth]{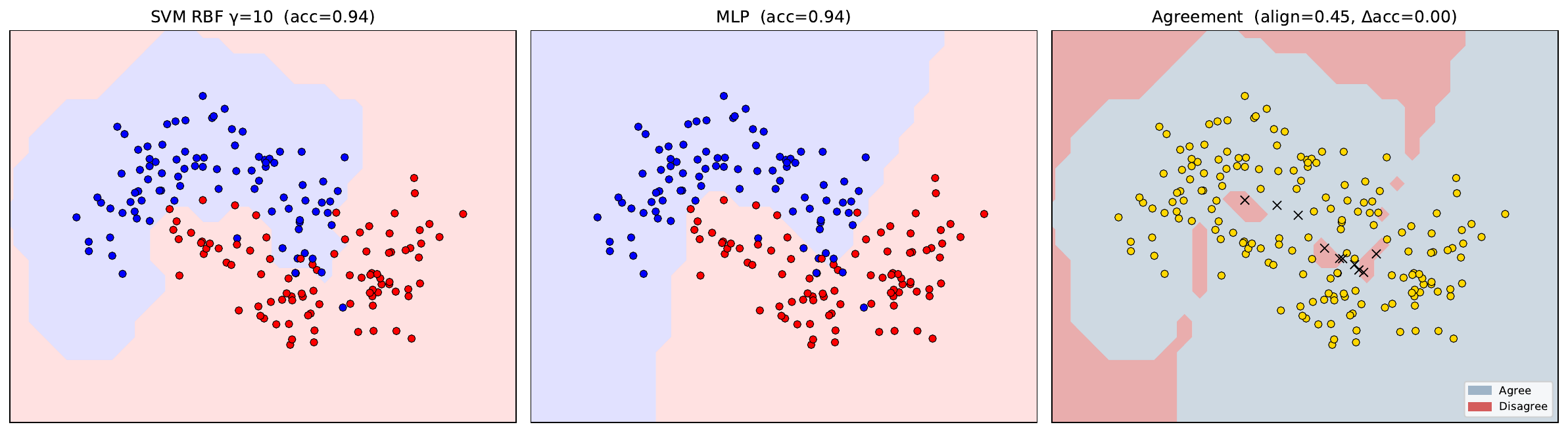}\\
  \includegraphics[width=0.85\textwidth]{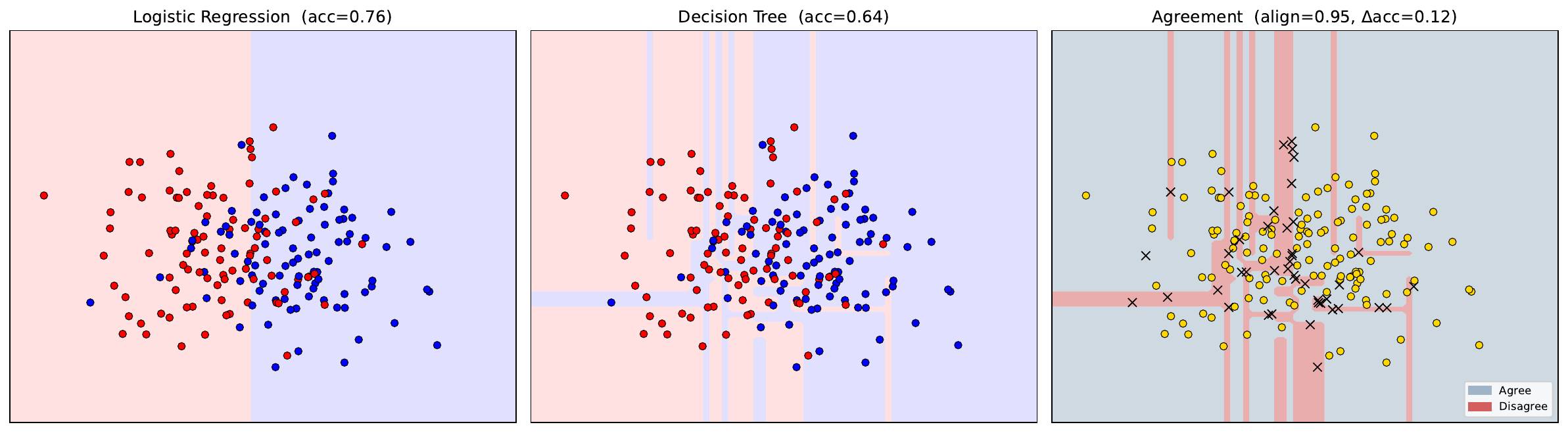}\\
  \includegraphics[width=0.85\textwidth]{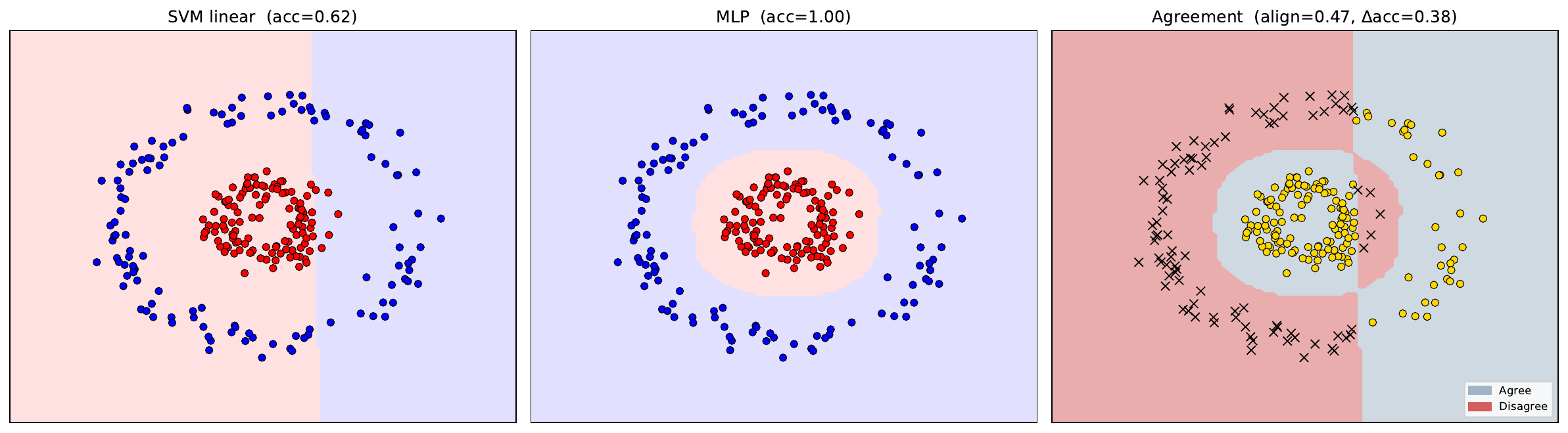}\\
  \includegraphics[width=0.85\textwidth]{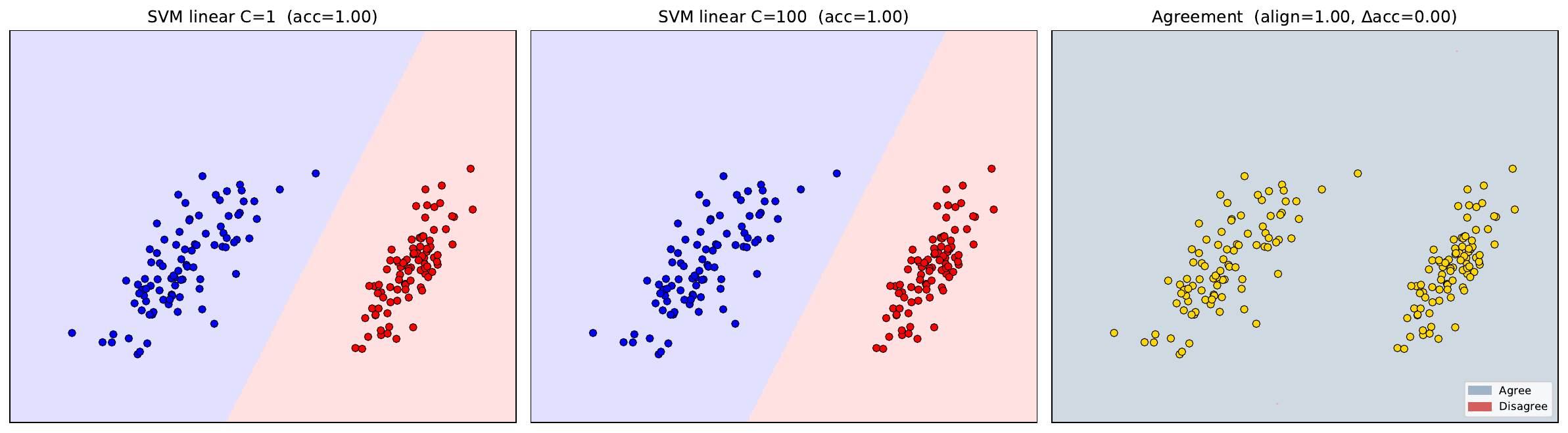}
  \caption{Four cases illustrating the relationship between gRA and accuracy difference. Each row shows two models (left, middle) and their agreement map (right): blue indicates agreement, red disagreement. Top to bottom: Case~1 (low gRA, low $\Delta$acc); Case~2 (high gRA, high $\Delta$acc); Case~3 (low gRA, high $\Delta$acc); Case~4 (high gRA, low $\Delta$acc).}
  \label{fig:cases}
\end{figure}

RA complements accuracy rather than replacing it. To make this concrete, we examine the four combinations of alignment (gRA) and accuracy difference between two models. Figure~\ref{fig:cases} illustrates each case on a synthetic 2D dataset, with the agreement map (right column) showing where the two models concur (blue) or disagree (red).

\textit{Case~1: Low gRA, low $\Delta$acc} (Fig.~\ref{fig:cases}, top row). An SVM with RBF kernel and an MLP both achieve accuracy $0.94$, yet agree on only $45\%$ of the instance space. Each model is accurate in different regions, and the test sample happens to lie where the errors balance out. Accuracy alone misses this structural divergence.

\textit{Case~2: High gRA, high $\Delta$acc} (Fig.~\ref{fig:cases}, second row). A logistic regression and a decision tree share nearly identical decision boundaries (gRA $= 0.95$) but differ by $12$ percentage points in accuracy. The disagreement is confined to thin axis-parallel bands induced by the tree, which happen to intersect a disproportionate share of test instances.

\textit{Case~3: Low gRA, high $\Delta$acc} (Fig.~\ref{fig:cases}, third row). A linear SVM (acc $= 0.62$) and an MLP (acc $= 1.00$) on a concentric-rings problem disagree on more than half the space. The MLP captures the non-linear structure that the linear SVM cannot, and gRA simply confirms what accuracy already reveals.

\textit{Case~4: High gRA, low $\Delta$acc} (Fig.~\ref{fig:cases}, bottom row). Two linear SVMs differing only in regularization ($C=1$ vs.\ $C=100$) achieve identical accuracy and gRA $= 1.00$. On a linearly separable problem, regularization strength does not change the learned boundary.

Cases~1 and~2 are where RA contributes the most: it surfaces structural similarities and differences that accuracy alone cannot.

\section{Experimental Setup}
\label{sec:setup}

To illustrate the usefulness of the proposed methodology, we compared pruned and unpruned decision trees on datasets from the UCI Machine Learning Repository~\cite{UCI2025}, using the Python \texttt{scikit-learn} library~\cite{pedregosa2011scikit} for model induction and the \texttt{ralign} package for the computation of dRA and gRA.%
\footnote{The \texttt{ralign} package and the experimental code are available at \url{https://github.com/mmrsantos/ralign}.} This scenario was chosen because the two models share the same hypothesis language, so their decision boundaries are directly comparable, while the well-known tendency of unpruned trees to overfit~\cite{Aggarwal2015} ensures that both highly aligned and strongly misaligned pairs can be found across different datasets, making the collection adequate to stress-test the measures.

\paragraph{Datasets.}
We used a diverse collection of 92 datasets from the UCI Machine Learning Repository~\cite{UCI2025}, covering multiple application domains (healthcare, finance, biology, image and signal processing), a wide range of sizes (32 to more than 4\,600 instances), dimensionalities (4 to 649 features), feature types (numerical, categorical, and mixed), and class distributions (balanced and imbalanced). The last column of each file was treated as the target; rows with a missing target and classes represented by fewer than three instances were discarded. Numerical features were imputed with the column median and standardised (zero mean, unit variance); categorical features were imputed with the most-frequent value and one-hot encoded.

\paragraph{Decision Tree Models.}
Both models were induced with \texttt{DecisionTreeClassifier}
(random state fixed at~42 throughout). The \emph{unpruned} tree was grown with \texttt{min\_samples\_split\,=\,2}, allowing splits down to single-sample nodes and producing the most complex tree the training data supports.

The \emph{pruned} tree used \texttt{min\_samples\_split\,=\,10} combined with cost-complexity pruning. The pruning procedure works as follows. Given a fully grown tree $T$, each internal node $t$ defines a subtree $T_t$ rooted at it. The \emph{effective complexity parameter} of that subtree is
\begin{equation}
  \alpha_{\mathrm{eff}}(t)
    = \frac{R(t) - R(T_t)}{\lvert T_t \rvert - 1},
  \label{eq:ccp}
\end{equation}
where $R(t)$ is the (impurity-weighted) error of node $t$ treated as a leaf, $R(T_t)$ is the sum of leaf errors within $T_t$, and $\lvert T_t \rvert$ is the number of leaves in $T_t$. Intuitively, $\alpha_{\mathrm{eff}}(t)$ measures the per-leaf reduction in training impurity that the subtree $T_t$ provides over simply declaring $t$ a leaf: a small value means the subtree adds many leaves for very little impurity gain and is therefore a good pruning candidate. Starting from the full tree, the node with the smallest $\alpha_{\mathrm{eff}}$ is pruned first, producing a simpler tree whose own $\alpha_{\mathrm{eff}}$ values are recomputed; the process is repeated until only the root remains. The result is a finite, monotonically non-decreasing sequence $0 = \alpha_0 \le \alpha_1 \le \cdots \le \alpha_K$, where each $\alpha_k$ is the threshold at which a new pruning event occurs~\cite{Aggarwal2015}.

The regularisation parameter $\alpha$ was selected per cross-validation fold. An inner 80/20 split of the fold's training data was used to compute this pruning path on the 80\,\% portion. The largest value in the sequence, $\alpha_K$, was then chosen, yielding the most aggressively pruned tree that the data supported, and that tree was re-trained on the full fold training partition with that $\alpha$.

\paragraph{Evaluation Protocol.}
Model performance and alignment were estimated using 5-fold cross-validation (shuffled, random seed 42). For each fold, both trees were trained on the training partition and evaluated on the held-out test partition. Three quantities were recorded per fold: (i)~accuracy difference (pruned minus unpruned), measuring whether pruning hurts or helps predictive performance; (ii)~dRA, computed on the test partition; and (iii)~gRA. Final per-dataset values were obtained by averaging the five fold estimates.

\paragraph{Synthetic Data for gRA.}
To measure gRA, 1\,000 synthetic instances were generated uniformly at random within the bounding box defined by the training features of each fold. No missing values were introduced in the synthetic data. The two trained models were applied to these instances and their agreement was computed as described in Sect.~\ref{sec:align}.

\paragraph{Visualisation.}
Results were analysed through three complementary views: (i)~histograms of the
distributions of gRA, dRA, and accuracy difference across datasets; (ii)~scatter plots of
gRA versus accuracy difference and of gRA versus dRA, each annotated with the
Pearson correlation coefficient; and (iii)~a quadrant scatter plot that partitions datasets
by the medians of gRA and $|\Delta\text{acc}|$, highlighting the four regimes described in Sect.~\ref{sec:disc-alig-vs-acc}.

\section{Results and Discussion}
\label{sec:results}

We carried out two sets of experiments to illustrate the usefulness of RA:

\begin{description}
  \item[E1:] comparison between the information obtained with gRA and accuracy difference;
  \item[E2:] comparison between the information obtained with gRA and dRA.
\end{description}

\subsection{E1: gRA and Accuracy Difference}

\begin{figure}[htbp]
  \centering
  \includegraphics[width=0.7\textwidth]{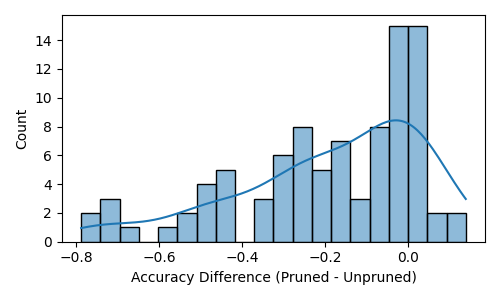}
  \caption{Distribution of accuracy difference (pruned minus unpruned) across 92
           datasets. The left-skewed shape indicates that pruning most often incurs
           an accuracy cost, though the magnitude varies widely.}
  \label{fig:hist_accdiff}
\end{figure}

Figure~\ref{fig:hist_accdiff} shows the distribution of the accuracy difference (pruned minus unpruned) across the 92 datasets. The distribution was strongly left-skewed: the most frequent outcome was a difference close to zero, but a long tail extended to values below $-0.8$, and only a small minority of datasets showed a positive difference. This pattern is consistent with the known tendency of unpruned trees to overfit~\cite{Aggarwal2015}: on datasets with complex or noisy boundaries, an unpruned tree memorises training noise and achieves spuriously high training accuracy, yet generalises poorly to unseen data, so that the pruned model can match or exceed it on the test set. On simpler datasets, however, pruning removes genuinely useful structure, which explains the long negative tail.

\begin{figure}[htbp]
  \centering
  \includegraphics[width=0.7\textwidth]{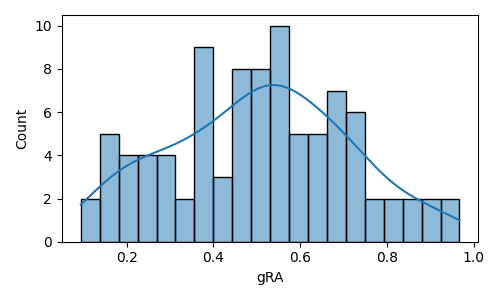}
  \caption{Distribution of gRA across 92 datasets. The spread across the full range
           reflects a wide variety of decision-boundary alignment between the two
           tree variants.}
  \label{fig:hist_gRA}
\end{figure}

Figure~\ref{fig:hist_gRA} shows the distribution of gRA. Unlike accuracy difference, gRA was spread across the full $[0, 1]$ range, with a broad peak between $0.4$ and $0.7$ and no strong concentration at either extreme. This spread indicates that geometric alignment is sensitive to structural properties of each dataset, capturing variation that does not reduce to a single value: even datasets with similarly small accuracy differences exhibited very different gRA values, suggesting that gRA encodes information that accuracy alone does not.

\begin{figure}[htbp]
  \centering
  \includegraphics[width=0.7\textwidth]{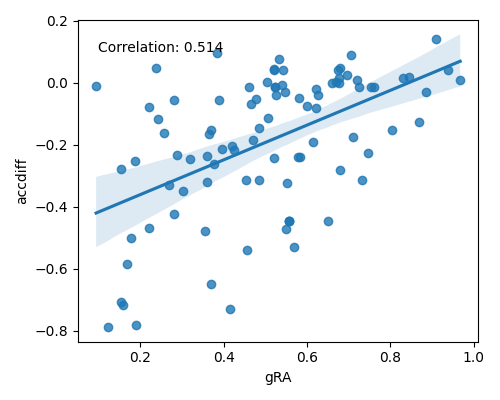}
  \caption{Scatter plot of gRA versus accuracy difference ($r = 0.514$). The positive
           trend indicates that higher geometric alignment is associated with smaller
           accuracy penalties, but the wide scatter shows that the two measures are
           complementary.}
  \label{fig:scatter_gRA_accdiff}
\end{figure}

\begin{figure}[htbp]
  \centering
  \includegraphics[width=0.75\textwidth]{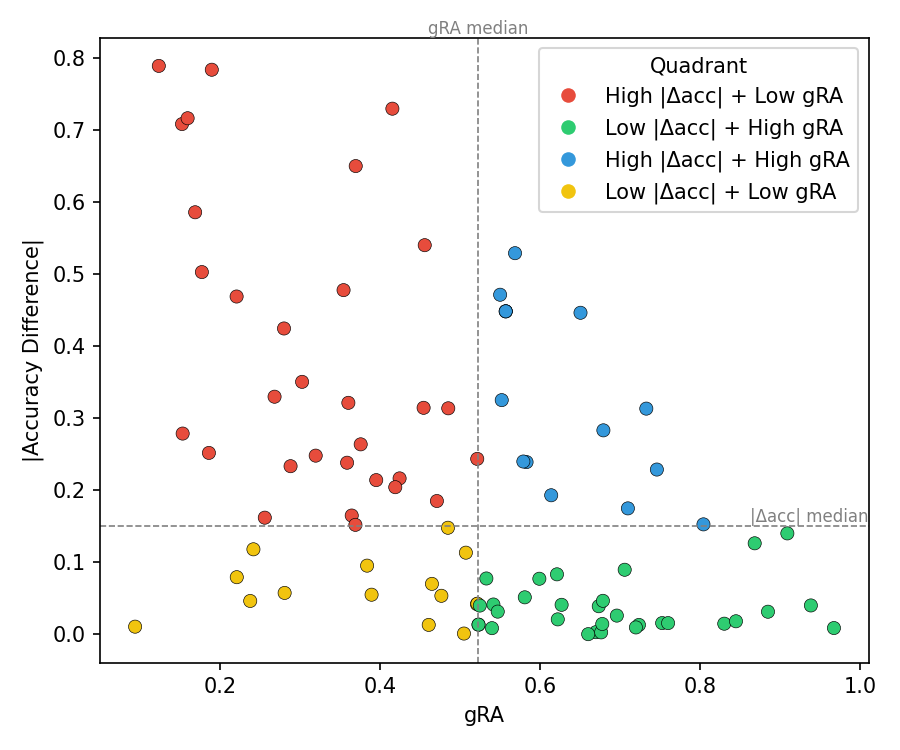}
  \caption{Quadrant analysis of gRA and $|\Delta\text{acc}|$ (dashed lines at their
           medians). The dominant quadrant is high $|\Delta\text{acc}|$ and low gRA
           (red). The blue quadrant identifies the critical case where high geometric
           alignment coexists with large accuracy differences.}
  \label{fig:quadrants_gRA_accdiff}
\end{figure}

Figure~\ref{fig:scatter_gRA_accdiff} shows the relationship between gRA and accuracy difference. A moderate positive correlation of $r = 0.514$ was observed, meaning that datasets where the two trees were geometrically similar tended to show smaller accuracy penalties from pruning. This is expected: if pruning does not change the decision boundary substantially across the instance space, it is unlikely to change test accuracy substantially either. However, the large scatter around the regression line confirms that the two measures are not redundant. Datasets with intermediate or high gRA could still produce large negative accuracy differences, and conversely some datasets with low gRA showed near-zero accuracy impact from pruning. These cases motivate the quadrant analysis in Fig.~\ref{fig:quadrants_gRA_accdiff}.

Figure~\ref{fig:quadrants_gRA_accdiff} partitions datasets by the medians of gRA ($\approx 0.52$) and $|\Delta\text{acc}|$ ($\approx 0.15$). The largest cluster was the \emph{high $|\Delta\text{acc}|$, low gRA} quadrant (red): on these datasets pruning changed both the decision boundary globally and test accuracy visibly. This is the expected outcome when the problem has complex boundaries and sufficient training data to grow a deep unpruned tree. The \emph{low $|\Delta\text{acc}|$, high gRA} quadrant (green) contained datasets where the two trees were both globally aligned and accuracy-equivalent, consistent with problems whose decision boundaries are simple and well-captured by either tree.

The two quadrants of particular interest are those where accuracy and geometric alignment are inconsistent. The \emph{high $|\Delta\text{acc}|$, high gRA} quadrant (blue) identified datasets where a large fraction of the instance space was classified identically by both trees, yet test accuracy differed substantially. This indicates that the disagreement between the models was concentrated in the region of the instance space covered by the test set, a configuration that geometric alignment alone would not flag. The \emph{low $|\Delta\text{acc}|$, low gRA} quadrant (yellow) was the least populated, which is consistent with the expectation that globally misaligned models are unlikely to produce equivalent test accuracy by chance. These results demonstrate that gRA and accuracy difference provide different and complementary perspectives on model similarity, and that neither measure alone is sufficient to characterise the relationship between two models.

\subsection{E2: gRA and Distributional RA}

\begin{figure}[htbp]
  \centering
  \includegraphics[width=0.7\textwidth]{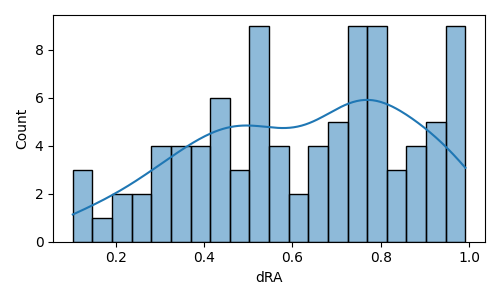}
  \caption{Distribution of dRA across 92 datasets. Values are concentrated towards
           higher agreement relative to gRA (Fig.~\ref{fig:hist_gRA}), indicating
           that models agree more on observed data than across the full instance space.}
  \label{fig:hist_dRA}
\end{figure}

Figure~\ref{fig:hist_dRA} shows the distribution of dRA. In contrast to gRA, dRA was concentrated towards higher values, with notable mass above $0.7$ and a secondary peak near $1.0$. This difference in location indicates that the pruned and unpruned trees tended to agree more on the observed test data than across the full instance space. Since the test set is a sample drawn from the data-generating distribution, dRA is higher in regions where training data is dense --- precisely those regions where both trees are likely to have learned the same boundary. Geometric alignment, by weighting all regions of the instance space equally, is thus a stricter criterion than distributional alignment.

\begin{figure}[htbp]
  \centering
  \includegraphics[width=0.7\textwidth]{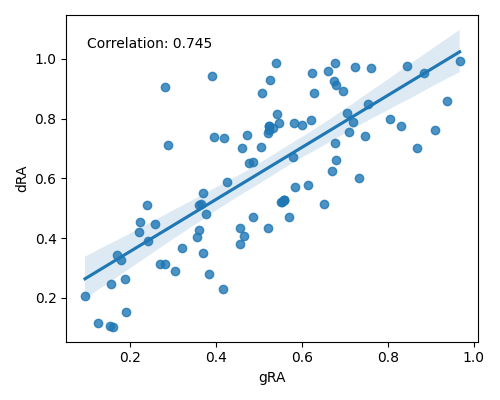}
  \caption{Scatter plot of gRA versus dRA ($r = 0.745$). Despite the strong overall
           trend, several datasets show high dRA with low gRA, identifying cases where
           distributional alignment overestimates global structural similarity.}
  \label{fig:scatter_gRA_dRA}
\end{figure}

Figure~\ref{fig:scatter_gRA_dRA} shows the relationship between gRA and dRA. A strong positive correlation of $r = 0.745$ was observed, confirming that the two measures capture related but distinct aspects of alignment. Notably, several datasets exhibited high dRA (above $0.8$) alongside low gRA (below $0.3$). These are cases where the two trees agreed on the test partition while differing substantially over most of the instance space, precisely the scenario illustrated in Fig.~\ref{fig:dif-dist-geom-align}. Relying solely on dRA in such cases would lead to an overestimate of structural model similarity. Conversely, no dataset showed high gRA with low dRA, which is consistent with the change-of-measure relationship between the two quantities derived in Sect.~\ref{sec:variants}: if two models agree across the whole space uniformly, they must also agree on any subsample of it, including the test set.

Taken together, these results suggest that gRA and dRA are complementary measures.
A high dRA alone is not sufficient to conclude that two models are structurally similar, as it may simply reflect the concentration of observed data in a region where the models happen to agree. Geometric alignment, by sampling uniformly from the instance space, provides a more complete and distribution-independent view of the similarity between decision boundaries.
 
\section{Conclusion}
\label{sec:conclusion}
 
In this paper, we introduced Geometric Rashomon Alignment (gRA), a measure to quantify the alignment between the decision boundaries defined by different models, using data uniformly sampled from the instance space defined by the attribute domains. Through extensive experiments on over 92 datasets, comparing pruned and unpruned decision trees, we show that RA complements the evaluation of predictive performance of models as well as their distributional alignment (i.e., computed on a sample of observed data).
Future work includes the use of RA to analyze other algorithms as well as the definition of other methods to sample from the instance space.

\section*{Acknowledgements}
This work is a result of Agenda “Center for Responsible AI”, nr. C645008882-00000055, investment project nr. 62, financed by the Recovery and Resilience Plan (PRR) and by European Union -  NextGeneration EU. Funded by the European Union – NextGenerationEU. Views and opinions expressed are however those of the author(s) only and do not necessarily reflect those of the European Union or the European Commission. Neither the European Union nor the European Commission can be held responsible for them. 
AISym4Med (101095387) supported by Horizon Europe Cluster 1: Health, ConnectedHealth (n.o 46858), supported by Competitiveness and Internationalisation Operational Programme (POCI) and Lisbon Regional Operational Programme (LISBOA 2020), under the PORTUGAL 2020 Partnership Agreement, through the European Regional Development Fund (ERDF). 
This work was financially supported by: UID/00027 of the LIACC - Artificial Intelligence and Computer Science Laboratory - funded by Fundação para a Ciência e a Tecnologia, I.P./ MCTES through the national funds.

\bibliographystyle{splncs04}
\bibliography{references}
 
\end{document}